# Boosting as a Product of Experts

**Narayanan U. Edakunni**
University of Bristol, UK

**Gavin Brown**
University of Manchester, UK

**Tim Kovacs**
University of Bristol, UK

## Abstract

In this paper, we derive a novel probabilistic model of boosting as a Product of Experts. We re-derive the boosting algorithm as a greedy incremental model selection procedure which ensures that addition of new experts to the ensemble does not decrease the likelihood of the data. These learning rules lead to a generic boosting algorithm - POEBoost which turns out to be similar to the AdaBoost algorithm under certain assumptions on the expert probabilities. The paper then extends the POEBoost algorithm to POEBoost.CS which handles hypothesis that produce probabilistic predictions. This new algorithm is shown to have better generalization performance compared to other state of the art algorithms.

## 1 Introduction

Boosting has been a popular form of ensemble classification that adaptively builds an ensemble of weak classifiers to achieve a good classification accuracy. Recently, research on boosting has focussed on linking the boosting procedure to various established machine learning paradigms like additive models [Friedman et al., 2000], minimization of Bregman distance [Collins et al., 2002], entropy projection [Kivinen and Warmuth, 1999] and gradient descent procedure [Mason et al., 1998]. These alternative frameworks for boosting provide new insights into its working which can then be used to extend and improve it. In this paper, we follow a similar line of research and provide the first ever probabilistic model of boosting. Here, we develop boosting as a Product of Experts (PoE) [Hinton, 2002] and derive the learning updates as a form of incremental model adaptation by adding new experts to the product. A probabilistic framework for boosting provides a number of advantages including a simple and well motivated model of the data. Furthermore, it makes the modeling assumptions made in boosting explicit and allows us to seamlessly apply boosting across different problem settings by varying the probabilistic model of the constituent experts. A probabilistic model of boosting also enables us to use a plethora of inference techniques like likelihood maximization and Bayesian inference to learn the parameters of the model.

In this paper, we model boosting as a normalized product of probabilities with the component probabilities being contributed by the experts in the ensemble. The ensemble of experts is expanded at each iteration by adding a new expert such that the likelihood of the observed data, as predicted by the ensemble, does not decrease with the addition of an expert. We show that such a condition of non-decreasing likelihood at each iteration naturally leads to a constraint on the parameters of the expert similar to that of a weak learning criterion in boosting [Freund and Schapire, 1997, Schapire, 1990]. For a specific parametrization of the expert probabilities we show that incremental learning in PoE reduces to a new variant of the boosting algorithm we call POEBoost. We demonstrate the value of a probabilistic model of boosting by deriving another new variant of boosting that uses base learners with probabilistic predictions as opposed to binary predictions. Using empirical evaluations, the new boosting algorithm is shown to have improved learning characteristics including better accuracy and correct levels of confidence in class predictions as compared to existing boosting procedures.

## 2 Boosting

We consider a problem of binary classification where we are given a training sample consisting of the input-class tuples $(x_1, y_1) \ldots (x_N, y_N)$ where each of the in-puts $x_i$ belongs to the domain $X$ and the class labels can take a value from $\{-1, 1\}$. We are interested in learning the model that relates input and corresponding class label and using this model to predict the class label of a previously unseen input $x_q$.

A boosting algorithm creates an ensemble of experts by incrementally adding *weak* hypotheses to the ensemble. Each successive weak hypothesis is chosen so as to minimize a weighted loss function of the data usually denoted as $\epsilon$. The boosting algorithm then assigns a weight $\alpha$ to the weak hypothesis that is proportional to its confidence of predicting the correct class label. The weights of the correctly classified data points are then decreased and that of the incorrect ones increased. These revised weights are then used to obtain the next weak classifier and the process repeats until convergence. There have been a number of variants of boosting that differ in the loss functions, the class of weak hypotheses and the way the weights are assigned. However, in this paper, we use AdaBoost as a popular instance of the family of boosting algorithms. The AdaBoost algorithm is illustrated in algorithm 1. In this paper, we establish a link between

---
**Algorithm 1** Learning in AdaBoost

Training data : $S = ((x_1, y_1) \ldots (x_N, y_N))$.
$D_i^1 = \frac{1}{N}$ for $i = 1 \ldots N$.
**for** j = 1 to M **do**
  Define $\epsilon_j = \sum_{i: h_j(x_i) \neq y_i} D_i^j$.
  Obtain a hypothesis $h_j$ that minimizes $\epsilon_j$ and satisfies the condition $\epsilon_j \leq 1/2$.
  $\alpha_j = \frac{1}{2} \log \left( \frac{1-\epsilon_j}{\epsilon_j} \right)$.
  $D_i^{j+1} = e^{(-y_i h_j(x_i) \alpha_j)} D_i^j$.
  $D_i^{j+1} = \frac{D_i^{j+1}}{\sum_i D_i^{j+1}}$.
**end for**
Prediction $H(x_q) = sign \left[ \sum_j \alpha_j h_j(x_q) \right]$.

---

an incremental learning strategy in PoE and boosting by showing that under certain parametrizations of the expert probabilities the learning algorithm of PoE has a structure identical to algorithm 1. We start by defining the framework of product of experts.

## 3 Product of Experts model

A Product of Experts model [Hinton, 2002] formulates the probability of a data point as a normalized product of probabilities with each probability being contributed by a different expert. When dealing with the problem of classification, we can express the probability of a class label conditioned on the input as a product of conditional class probabilities as shown:

$$P(Y = y | X = x, h_1 \ldots h_M) =$$
$$\frac{\prod_j^M P(Y = y | X = x, h_j)}{\prod_j^M P(Y = y | X = x, h_j) + \prod_j^M P(Y = \overline{y} | X = x, h_j)} \quad (1)$$

where $X$ is the random variable corresponding to the input, $x$ is a particular realization of $X$, $Y$ the random variable corresponding to the response, $y$ its realization, $\overline{y}$ the complement class of $y$ and $h_1 \ldots h_M$ the experts. We find from eq. (1) that the *ensemble probability* $P(Y = y | X = x, h_1 \ldots h_M)$ is expressed as a normalized product of the *expert probabilities* $P(Y = y | X = x, h_j)$ such that $P(Y = y | X = x, h_1 \ldots h_M) + P(Y = \overline{y} | X = x, h_1 \ldots h_M) = 1$. We can now recursively define the probability of an ensemble in terms of the probability of the current expert and the probability of the previous ensemble as

$$P(y|x, h_1 \ldots h_M) = P(y|x, h_M) P(y|x, h_1 \ldots h_{M-1}) / $$
$$[P(y|x, h_M) P(y|x, h_1 \ldots h_{M-1}) + $$
$$P(\overline{y}|x, h_M) P(\overline{y}|x, h_1 \ldots h_{M-1})] \quad (2)$$

where we have dropped the random variables from the formulation for notational convenience. When we add a new expert, the recursive definition of eq. (2) allows us to update the probability of an existing ensemble using the probability of a newly added expert. In later sections, we will use the recursive definition to develop routines for incremental expansion of the ensemble with simple update rules for the probability of the ensemble.

The probabilistic framework of PoE offers a simple model of the observed data, in which the probability of a data point is obtained by combining the opinions of a set of experts. PoE does not specify how the experts are generated, although there are principled probabilistic tools to fit the experts to the observed data.

Our aim is to use PoE to model the boosting approach to learning, where experts (weak learners) are generated incrementally. AdaBoost, for example, adds an expert which minimizes the error function in algorithm 1. Here, we will add experts to PoE incrementally, using a probabilistic approach. This approach views the PoE as a model of the observed data and hence new experts should be selected to improve the fit of the PoE to the data. The criteria for judging the fit is the likelihood of the observed data conditioned on the PoE model. Hence we would like to obtain an iterative procedure that increases the likelihood of the data with every addition of an expert.

## 3.1 Constraint on an expert

We next derive the condition on a newly added expert that would ensure that the likelihood of the observed data does not decrease with the addition of a new expert to the ensemble. We then refine this condition to show that this constraint on the expert leads to a weak learning condition on the expert akin to the one derived in [Freund and Schapire, 1997] for boosting, thus establishing part of the relation between PoE and boosting.

We start with the conditional likelihood of the IID data given by

$$P(y_1 \ldots y_N | x_1 \ldots x_N, h_1 \ldots h_j) = \prod_{i=1}^{N} P(y_i | x_i, h_1 \ldots h_j). \quad (3)$$

The corresponding likelihood for a PoE with $j-1$ experts is given by

$$P(y_1 \ldots y_N | x_1 \ldots x_N, h_1 \ldots h_{j-1}) = \prod_{i=1}^{N} P(y_i | x_i, h_1 \ldots h_{j-1}). \quad (4)$$

With every addition of an expert we would like the likelihood of the observed data, conditioned on the ensemble, to increase or remain the same. This condition on the likelihood can be expressed mathematically as

$$P(y_1 \ldots y_N | x_1 \ldots x_N, h_1 \ldots h_j) \geq P(y_1 \ldots y_N | x_1 \ldots x_N, h_1 \ldots h_{j-1}). \quad (5)$$

The inequality in eq. (5) leads to a constraint on the probabilities of the added expert (refer to appendix A for the derivation) expressed as

$$\sum_{i=1}^{N} \frac{D_i^{j-1}}{P(y_i | x_i, h_j)} \leq 2 \quad (6)$$

where $D_i^{j-1} = \frac{P(\bar{y}_i | x_i, h_1 \ldots h_{j-1})}{\sum_{i=1}^{N} P(\bar{y}_i | x_i, h_1 \ldots h_{j-1})}$ such that $\sum_i D_i^{j-1} = 1$. Examining the constraint in eq. (6), we find that the constraint is expressed as a bound on the weighted sum of the reciprocals of the probabilities of the observations as assigned by the newly added expert. The weight $D_i^{j-1}$, in turn, is proportional to the probability that the ensemble with $j-1$ experts made a mistake in predicting the $i^{th}$ data point.

We note that if the constraint in eq. (6) is valid, adding experts iteratively results in the guarantee of a non-decreasing likelihood. Furthermore, weights $D_i^j$ can be computed from the previous weights $D_i^{j-1}$ in each iteration by using the recursive definition of PoE given in eq. (2). We start with the definition of $D_i^j$:

$$D_i^j = \frac{P(\bar{y}_i | x_i, h_1 \ldots h_j)}{\sum_{i=1}^{N} P(\bar{y}_i | x_i, h_1 \ldots h_j)}. \quad (7)$$

Using eq. (2) we can simplify eq. (7) into

$$\begin{aligned} D_i^j &= \frac{P(\bar{y}_i | x_i, h_j) P(\bar{y}_i | x_i, h_1 \ldots h_{j-1}) / Q_i^j}{\sum_{i=1}^{N} P(\bar{y}_i | x_i, h_j) P(\bar{y}_i | x_i, h_1 \ldots h_{j-1}) / Q_i^j} \\ &= \frac{P(\bar{y}_i | x_i, h_j) D_i^{j-1} / Q_i^j}{\sum_{i=1}^{N} P(\bar{y}_i | x_i, h_j) D_i^{j-1} / Q_i^j} \end{aligned} \quad (8)$$

where $Q_i^j$ is the normalization term given by

$$\begin{aligned} Q_i^j &= P(y_i | x_i, h_j) P(y_i | x_i, h_1 \ldots h_{j-1}) \\ &+ P(\bar{y}_i | x_i, h_j) P(\bar{y}_i | x, h_1 \ldots h_{j-1}). \end{aligned} \quad (9)$$

We can now iteratively add an expert to the ensemble satisfying the constraint given in eq. (6) and update the weights on the data as given by eq. (8). This results in a procedure as illustrated in algorithm 2. Algorithm

---
**Algorithm 2** Incremental learning in PoE

Training data : $S = ((x_1, y_1) \ldots (x_N, y_N))$
$D_i^1 = 1/N$ for $i = 1 \ldots N$.
**for** j = 1 to M **do**
  Obtain a hypothesis $h_j$ such that $\sum_{i=1}^{N} \frac{D_i^j}{P(y_i | x_i, h_j)} \leq 2$
  $D_i^{j+1} = P(\bar{y}_i | x_i, h_j) D_i^j$
  $D_i^{j+1} = \frac{D_i^{j+1}}{\sum_i D_i^{j+1}}$
**end for**
$H(x_q) = sign\,[P(Y_q = 1 | x_q, h_1 \ldots h_M)$
$- P(Y_q = -1 | x_q, h_1 \ldots h_M)].$

---

2 is broadly similar to algorithm 1 in its structure but differs significantly in the details. In the following sections, we show that the boosting algorithm as exemplified by algorithm 1 is just a specific case of the generic algorithm given in algorithm 2.

## 4 Boosting as incremental learning in PoE

We can derive the boosting updates given in algorithm 1 from the incremental learning that we had formulated in the previous section by assuming a particular parametrization for the expert probabilities. In this section, we start with a specific parametrization for the expert probability and derive the weak learning criteria as a specific instance of the constraint given in eq. (6). We then derive the values for the weights $\alpha$ on the hypothesis and the update rules for the weights on the data $D_i$ used in boosting.

Let us assume that the conditional probability contributed by an expert is

$$\begin{aligned} P(Y = y | X &= x, h(x)) = \\ P(Y &= y | Z = y) P(Z = y | x, h) + \\ P(Y &= y | Z = \bar{y}) P(Z = \bar{y} | x, h) \end{aligned} \quad (10)$$

where $Z$ is a random variable representing the prediction made by hypothesis $h$ and hence takes values from $\{-1, 1\}$. In eq. (10), the probability of an expert predicting a particular label is given as a weighted combination of the probabilities of the predictions of $h$. The weights $P(Y = y|Z = y)$ and $P(Y = y|Z = \bar{y})$ are the probabilities that the observed class label is the same as the one predicted by the hypothesis or different from the prediction of the hypothesis. These probabilities can thus be interpreted as an error probability associated with the hypothesis that is independent of the input and distorts the predictions of the hypothesis. We further assume that the distortions are symmetric resulting in the expression

$$P(Y = -1|Z = 1) = P(Y = 1|Z = -1) = P_e \quad (11)$$

where we have represented the error probabilities using a constant parameter $P_e$ to signify that it is independent of the input and is a parameter associated with a particular hypothesis. Furthermore, to derive the weak learning condition we also assume that the hypothesis predicts with perfect confidence :

$$P(Z = y|x, h), P(Z = \bar{y}|x, h) \in \{0, 1\}. \quad (12)$$

Substituting the error probability given in eq. (11) in eq. (10) we obtain

$$P(Y = y|X = x, h(x)) = $$
$$(1 - P_e)(1 - P(Z = \bar{y}|x, h)) + P_e P(Z = \bar{y}|x, h). \quad (13)$$

We can now substitute the expert probability as given by eq. (13) into the constraint for expert $j$ as given in eq. (6) to get

$$\sum_{i=1}^{N} \frac{D_i^{j-1}}{(1 - P_e^j)(1 - P(Z = \bar{y}_i|x, h_j)) + P_e^j P(Z = \bar{y}_i|x, h_j)} \leq 2 \quad (14)$$

where $P_e^j$ is the error parameter of the $j^{th}$ expert. We note that $P(\bar{y}_i|x, h) = 1$ for all the data points where $h(x_i) \neq y_i$ and $P(\bar{y}_i|x, h) = 0$ for the data points where $h(x_i) = y_i$. Using this condition, we can simplify eq. (14) as

$$\sum_{i:h_j(x_i)=y_i} \frac{D_i^{j-1}}{(1 - P_e^j)} + \sum_{i:h_j(x_i)\neq y_i} \frac{D_i^{j-1}}{P_e^j} \leq 2. \quad (15)$$

Making use of the relation $\sum_{i=1}^{N} D_i^{j-1} = 1$ we can simplify eq. (15) to obtain the constraint on $P_e^j$ as

$$\epsilon_j \leq P_e^j \leq \frac{1}{2} \quad (16)$$

where $\epsilon_j = \sum_{i:h_j(x_i)\neq y_i} D_i^{j-1}$. The constraint in eq. (16) implies $\epsilon_j \leq \frac{1}{2}$ which translates to a constraint on the base hypothesis of an expert. The base hypothesis chosen at each iteration of the algorithm must be such that the sum of weights of the data points on which it makes an error in prediction is less than $1/2$. This constraint is the popular weak learning criterion of boosting as formulated in [Freund and Schapire, 1997]. Hence, we started with the generic constraint on the expert of a PoE as defined in eq. (6) and derived a constraint on the base hypothesis when the probability of an expert is parametrized as in eq. (10) and satisfies assumptions given by eq. (11) and eq. (12). This demonstrates one part of the similarity between PoE and boosting.

In the next step, we reparametrize $P_e$ to derive the weights on the base hypothesis $\alpha$ and the updates for the weights $D$ on the data points. We previously derived a constraint on $P_e$ in eq. (16) which allows $P_e$ to take a range of values from $\epsilon$ to $1/2$. We can be greedy about our choice of $P_e$ and assign maximum confidence to our expert by choosing the minimum value for $P_e$. Hence, in our algorithm we choose $P_e^j = \epsilon_j$. Now, we can reparametrize $P_e^j$ as

$$P_e^j = \frac{e^{-\alpha_j}}{(e^{-\alpha_j} + e^{\alpha_j})}. \quad (17)$$

Setting $P_e^j = \epsilon_j$ in eq. (17) and solving for $\alpha_j$ we get

$$\alpha_j = \frac{1}{2} \log \frac{1 - \epsilon_j}{\epsilon_j}. \quad (18)$$

Next, we derive the updates for the data points $D_i^{j-1}$. The update for $D_i$ as given in eq. (8) can be modified to our specific model by plugging the value of $P(\bar{y}_i|x_i, h_j)$ in eq. (8). The value of $P(\bar{y}_i|x_i, h_j)$ can be computed from eq. (13) as

$$P(\bar{y}_i|x_i, h_j) = $$
$$P_e^j(1 - P(Z = \bar{y}_i|x_i, h_j)) + (1 - P_e^j)P(Z = \bar{y}_i|x_i, h_j). \quad (19)$$

Using the assumptions in eq. (12) about $P(Z = \bar{y}_i|x_i, h_j)$ we can rewrite $P(\bar{y}_i|x_i, h_j)$ as

$$P(\bar{y}_i|x_i, h_j) = \begin{cases} P_e^j & \text{if } h_j(x_i) = y_i \\ (1 - P_e^j) & \text{if } h_j(x_i) \neq y_i. \end{cases} \quad (20)$$

Substituting the value of $P_e^j$ as given by eq. (17) we obtain

$$P(\bar{y}_i|x_i, h_j) = \begin{cases} \frac{e^{-\alpha_j}}{(e^{-\alpha_j} + e^{\alpha_j})} & \text{if } h_j(x_i) = y_i \\ \frac{e^{\alpha_j}}{(e^{-\alpha_j} + e^{\alpha_j})} & \text{if } h_j(x_i) \neq y_i \end{cases}$$
$$= \frac{e^{(-y_i h_j(x_i) \alpha_j)}}{(e^{-\alpha_j} + e^{\alpha_j})} \quad (21)$$

where we have used the relation $y_i, h_j(x_i) \in \{-1, 1\}$ to simplify the expression for $P(\overline{y}_i|x_i, h_j)$. Correspondingly, $P(y_i|x_i, h_j)$ is given by

$$P(y_i|x_i, h_j) = 1 - P(\overline{y}_i|x_i, h_j) = \frac{e^{(y_i h_j(x_i)\alpha_j)}}{(e^{-\alpha_j} + e^{\alpha_j})} \quad (22)$$

We can now use the value of $P(\overline{y}_i|x_i, h_j)$ to update the weights $D_i^{j-1}$ using eq. (8) as :

$$D_i^j = \frac{e^{(-y_i h_j(x_i)\alpha_j)} D_i^{j-1}/Q_i^j}{\sum_{i=1}^N e^{(-y_i h_j(x_i)\alpha_j)} D_i^{j-1}/Q_i^j} \quad (23)$$

where $Q_i^j = e^{(y_i h_j(x_i)\alpha_j)} P(y_i|x_i, h_1 \ldots h_{j-1}) + e^{(-y_i h_j(x_i)\alpha_j)} P(\overline{y}_i|x, h_1 \ldots h_{j-1})$.

In the final step we derive the rules of prediction for the boosting procedure that we have developed. The prediction in a PoE is based on the probability assigned to a class label given the test input $x_q$. The probabilities of different class labels are given by

$$P(Y_q = 1|X_q = x_q, h_1 \ldots h_M) = \frac{\prod_{j=1}^M P(1|x_q, h_j)}{R_q}$$
$$P(Y_q = -1|X_q = x_q, h_1 \ldots h_M) = \frac{\prod_{j=1}^M P(-1|x_q, h_j)}{R_q} \quad (24)$$

where $R_q = \prod_{j=1}^M P(1|x_q, h_j) + \prod_{j=1}^M P(-1|x_q, h_j)$. The class label for an input is decided by the greater of the two probabilities given in eq. (24). We can alternatively express the class label $y_q$ corresponding to $x_q$ as

$$\begin{aligned} y_q = sign\,[&P(Y_q = 1|X_q = x_q, h_1 \ldots h_M) \\ &- P(Y_q = -1|X_q = x_q, h_1 \ldots h_M)]. \end{aligned} \quad (25)$$

We can substitute the values of the expert probabilities as given in eq. (22) and eq. (24) in eq. (25) to obtain

$$\begin{aligned} y_q &= sign\left[\prod_j \frac{e^{(h_j(x_i)\alpha_j)}}{(e^{-\alpha_j} + e^{\alpha_j})} - \prod_j \frac{e^{(-h_j(x_i)\alpha_j)}}{(e^{-\alpha_j} + e^{\alpha_j})}\right] \\ &= sign\left[e^{\left(\sum_j \alpha_j h_j(x_i)\right)} - e^{\left(-\sum_j \alpha_j h_j(x_i)\right)}\right] \\ &= sign\left[\sum_j \alpha_j h_j(x_i)\right]. \end{aligned} \quad (26)$$

The updates derived for the boosting algorithm have been summarized in algorithm 3 which we name as the *POEBoost.DS* algorithm where DS stands for the **D**iscrete nature of the predictions made by the base hypothesis and **S**ymmetric error probabilities $P_e$. We find that POEBoost.DS is similar to AdaBoost to a large extent including the weak learning constraint on the hypothesis and the learning rules for $\alpha$. The only difference between the two algorithms lie in the way that $D$ gets updated.

---

**Algorithm 3** Learning POEBoost.DS

Training data : $S = ((x_1, y_1) \ldots (x_N, y_N))$.
$D_i^1 = \frac{1}{N}$.
**for** j = 1 to M **do**
  Define $\epsilon_j = \sum_{i:h_j(x_i)\neq y_i} D_i^j$.
  Obtain a hypothesis $h_j$ that minimizes $\epsilon_j$ and satisfies the condition $\epsilon_j \leq 1/2$.
  $\alpha_j = \frac{1}{2} \log\left(\frac{1-\epsilon_j}{\epsilon_j}\right)$.
  $D_i^{j+1} \propto e^{(-y_i h_j(x_i)\alpha_j)} D_i^j$ (refer eq. (23)).
  $D_i^{j+1} = \frac{D_i^{j+1}}{\sum_i D_i^{j+1}}$.
**end for**
Prediction $H(x_q) = sign\left[\sum_j \alpha_j h_j(x_q)\right]$.

---

## 5 POEBoost.CS

In this paper, we had derived a boosting procedure with PoE as the underlying probabilistic model. The derived boosting procedure, under certain assumptions, evolved into a procedure which is remarkably similar to AdaBoost. There are deeper connections between POEBoost.DS and other conventional boosting techniques like AdaBoost and LogitBoost but to explore the entire depth of the relations is not possible within the scope of this paper.

Establishing a probabilistic model for boosting and deriving a basic boosting algorithm like POEBoost.DS provides new insights into the working of boosting and provides a statistical tool in the form of the PoE model to study the system in detail. However, the effectiveness of such a generic probabilistic model of PoE and its associated incremental learning rules can be best demonstrated by extending it to different learning problems. As a case study, we extend the PoE model to a scenario where the base hypothesis produces a continuous value for prediction, specifically we will deal with the case where the base hypothesis provides a probabilistic prediction of the class label. Using this case study we would like to demonstrate the ease with which such models can be derived from the basic PoE model and the learning rules formulated. We name this algorithm as POEBoost.CS, standing for **C**ontinuous predictions of the hypothesis and **S**ymmetric error probability.

In this section, we proceed in the same manner as we derived the boosting algorithm for POEBoost.DS - by deriving the weak learning assumption and using that to learn the parameters of the base hypothesis and the error probability. Real AdaBoost [Schapire and Singer, 1998] is the extension of AdaBoost that handles hypotheses with real valued predictions. However, this algorithm lacks a proper motivation other than that of an extension to handle real valued predictions. It needs alternate techniques to handle multiple labels and multiclass classification which is naturally compatible with the framework of PoE. We also demonstrate that POEBoost.CS is competitive in its generalization ability when compared to Real AdaBoost on benchmark datasets.

### 5.1 Constraint on the expert

We start by deriving the constraint on the expert under the conditions :

$$P(Z = y|x, h), P(Z = \bar{y}|x, h) \in [0, 1] \quad (27)$$
$$P(Y = -1|Z = 1) = P(Y = 1|Z = -1) = P_e. \quad (28)$$

Using these conditions, we reformulate the constraint given in eq. (6) as

$$\sum_{i=1}^{N} \frac{D_i^{j-1}}{P(y_i|x_i, h_j)} \leq 2 \quad (29)$$

where $P(y_i|x_i, h_j) = (1 - P_e)(1 - P(Z = \bar{y}|x, h)) + P_e P(Z = \bar{y}|x, h)$. We can substitute the value of $P(y_i|x_i, h_j)$ in the inequality of eq. (29), but the continuous values taken by the predictions of the base hypothesis prevent us from obtaining a closed form constraint for $P_e$. We can, however, relax the constraint by finding a piecewise linear upper bound for the expression in eq. (29) and constrain it to be less than 2 (refer to appendix B for details). The modified bound is then given by

$$\sum_{i=1}^{N} \frac{D_i^{j-1}}{P(y_i|x_i, h_j)}$$
$$\leq \sum_{i:\mathcal{C}_1} D_i^{j-1} \left[ 2 \left( 2 - \frac{1}{P_e} \right) P(Z = y_i|x_i, h_j) + \frac{1}{P_e} \right]$$
$$+ \sum_{i:\mathcal{C}_2} D_i^{j-1} \left[ 2 \left( \frac{1}{1 - P_e} - 2 \right) P(Z = y_i|x_i, h_j) \right.$$
$$+ \left. \left( 4 - \frac{1}{1 - P_e} \right) \right] \leq 2. \quad (30)$$

where $i : \mathcal{C}$ denotes the set of all indices $i$ that satisfy the condition given by $\mathcal{C}$. In eq. (30) the conditions $\mathcal{C}_1$ and $\mathcal{C}_2$ are given by $P(Z = y_i|x_i, h_j) \leq 0.5$ and $P(Z = y_i|x_i, h_j) > 0.5$ respectively. Now, we can solve for $P_e$ to obtain

$$\epsilon_j^c \leq P_e \leq \frac{1}{2} \quad (31)$$

where $\epsilon_j^c$ is given by

$$\epsilon_j^c = \left[ \sum_{i:\mathcal{C}_1} D_i^{j-1} \left( 2P(Z = y_i|x_i, h_j) - 1 \right) \right] /$$
$$\left[ 2 \sum_{i:\mathcal{C}_1} D_i^{j-1} \left( P(Z = y_i|x_i, h_j) - 1 \right) \right. \quad (32)$$
$$\left. -2 \sum_{i:\mathcal{C}_2} D_i^{j-1} P(Z = y_i|x_i, h_j) + 1 \right].$$

Table 1: Details of datasets used. The letters dataset was converted to a binary classification by only using data corresponding to letters A and B.

| Name | # data pts. | # features |
|---|---|---|
| Ionosphere | 351 | 33 |
| Spambase | 4601 | 56 |
| Breast cancer | 569 | 30 |
| Pima Indians diabetes | 768 | 8 |
| Letter | 1555 | 16 |
| Wine quality | 1599 | 11 |
| Catalysis (Train) | 873 | 617 |
| Catalysis (Test) | 300 | 617 |

The optimal values for $P_e^j$ and $h_j$ are chosen such that $P_e$ is minimized. The settings for which $P_e^j$ is minimized are $P_e^j = \epsilon_j^c$ and $h_j = argmin_h(\epsilon_j^c)$. The update for $D$ is given by eq. (8) and the prediction rule by eq. (26). This set of learning rules defines the POEBoost.CS boosting algorithm. We next compare the efficacy of this algorithm with that of related algorithms in the rest of the paper.

## 6 Evaluation

In this section, we evaluate the generalization performance of POEBoost.CS with that of Real AdaBoost [Schapire and Singer, 1998]. The POEBoost.CS[1] assumes that the base hypothesis predicts with the probability of a class rather than just the prediction of the class label. Other boosting algorithms that minimize a loss function of negative log likelihood like LogitBoost [Friedman et al., 2000] or GBM [Friedman, 2001] can also handle base classifiers with probabilistic predictions. However, these algorithms are designed to minimize a squared loss thus requiring a regression algorithm as a base classifier. This is different from the requirements for the base classifier of our boosting algorithm and hence we do not include the logistic regression based boosting algorithms in our comparison.

In typical boosting algorithms, decision stumps are used as base classifiers. Decision stumps can provide probabilistic predictions but they are typically domain partitioning algorithms that assign the same probability to all points lying to one side of the decision boundary. However, in more realistic probabilistic classifiers different data points are assigned different confidence levels depending on their distance from the decision boundary. Hence, we use a univariate logistic regressor as our base classifier which assigns probability of a class to an input depending on its distance from the decision boundary. At every round of hypothesis selection, a logistic regression is trained on an input fea-ture such that it minimizes a certain error criteria. At each iteration, the feature with the least error is chosen to build the logistic regression. We learn the parameters of the logistic regression by using a single step gradient descent on the error. We use the same algorithms and the same settings for the base classifiers for both the boosting algorithms to make a fair comparison. The only difference between the base classifiers used in these algorithms are the criteria that is used to measure the error. The algorithm of POEBoost.CS uses $\epsilon^c$ as the error which is minimized to choose the best base classifier whereas Real AdaBoost uses $-\sum_i D_i P(y_i|x_i, h_j)$ as the error to minimize [Schapire and Singer, 1998].

---
[1]Code available at http://www.cs.bris.ac.uk/~nara/

Table 2: Comparison of boosting algorithms with figures in bold indicating better performance.

| Name | Real AdaBoost | | POEBoost.CS | |
|---|---|---|---|---|
| | Test accuracy | Test likelihood | Test Accuracy | Test likelihood |
| Ionosphere | 0.78(0.03) | -0.52(0.05) | **0.85**(0.03) | **-0.41**(0.05) |
| Breast cancer | 0.84(0.04) | -0.51(0.02) | **0.96**(0.02) | **-0.12**(0.03) |
| Spambase | 0.78(0.01) | -0.65(0.00) | **0.86**(0.01) | **-0.39**(0.01) |
| Pima Indians diabetes | 0.71(0.03) | -0.63(0.02) | **0.73**(0.03) | **-0.56**(0.03) |
| Letter | 0.54(0.02) | -0.59(0.02) | **0.94**(0.01) | **-0.15**(0.02) |
| Wine quality | 0.96(0.01) | -0.16(0.02) | 0.96(0.01) | -0.16(0.02) |
| Catalysis | 0.63 | -0.64 | **0.68** | **-0.59** |

We compare the two algorithms on the UCI datasets and the *catalysis* dataset obtained from the predictive uncertainty challenge.[2] The details of the data are given in table 1. The *catalysis* dataset consists of distinct training and validation data. Here we used the validation data to measure the test error due to the unavailability of the actual test data for *catalysis*. For the UCI datasets, the results were obtained after running the algorithms on 10 train-test splits with 75% of the data used for training and the rest for testing. The results reported are the average of these 10 experiments. In all these experiments 200 base classifiers were used during the learning phase. The test error and the likelihood are given in table 2.

The results in table 2 demonstrate the efficacy of the POEBoost.CS algorithm with respect to the state of the art boosting algorithm for confidence-rated predictions. The POEBoost.CS algorithm is seen to predict accurately with optimal confidences than its competitor in all of the data sets except in the *wine* dataset where both the algorithms show equal accuracy and confidence. The superiority of POEBoost.CS arises from the fact that boosting is framed as a probabilistic model which allows us to formulate a tight bound for the increase in the likelihood and maximizing this bound leads to a model that fits the data better.

---

[2]http://predict.kyb.tuebingen.mpg.de/pages/home.php

## 7 Discussion

We have presented a novel interpretation of boosting algorithms as a Product of Experts probabilistic model. The sequential distribution updates in the boosting procedure were explained as an iterative model selection process, adding experts such that the conditional likelihood is increased at each step. The model we have presented can incorporate arbitrary probabilistic models as the experts, including real-valued predictions naturally. It can also handle learning problems other than classification by choosing the appropriate probability model for the experts. The updates are derived naturally from a constraint on the likelihood (eq. (6)), defining a family of algorithms that satisfy this condition. In a particular special case, and when using experts of the form in eq. (10), the model reduces to an AdaBoost-like algorithm, with the only difference being that the weights on data points used in POEBoost are normalized across the classes whereas for AdaBoost they are not. Future work will focus on how to determine good settings for $\alpha$ in a given range, and analyzing the rates of convergence in these situations.

# 8 Appendix

## 8.1 Appendix A

In this appendix, we derive the constraint that ensures a non-decreasing likelihood for every addition of an expert to the ensemble. The log likelihood for an ensemble with $j$ experts is given by

$$\mathcal{L}_j = \log P(y_1 \ldots y_N | x_1 \ldots x_N, h_1 \ldots h_j) \qquad (33)$$

and for $j-1$ experts it is given by

$$\mathcal{L}_{j-1} = \log P(y_1 \ldots y_N | x_1 \ldots x_N, h_1 \ldots h_{j-1}). \qquad (34)$$

The requirement that $\mathcal{L}_j \geq \mathcal{L}_{j-1}$ results in

$$\sum_i \log \left( Q_{j-1}^i + \frac{P(\overline{y}_i|x_i, h_j)}{P(y_i|x_i, h_j)} \overline{Q}_{j-1}^i \right) \leq 0 \qquad (35)$$

where $Q_{j-1}^i = P(y_i|x_i, h_1 \ldots h_{j-1})$ and $\overline{Q}_{j-1}^i = P(\overline{y}_i|x_i, h_1 \ldots h_{j-1})$. We cannot find a closed form solution for the inequality in eq. (35) and hence we relax the bound using Jensen's inequality

$$\begin{aligned}
&\frac{1}{N} \sum_i \log \left( Q_{j-1}^i + \frac{P(\overline{y}_i|x_i, h_j)}{P(y_i|x_i, h_j)} \overline{Q}_{j-1}^i \right) \\
&\leq \log \frac{1}{N} \sum_i \left( Q_{j-1}^i + \frac{P(\overline{y}_i|x_i, h_j)}{P(y_i|x_i, h_j)} \overline{Q}_{j-1}^i \right) \leq 0.
\end{aligned} \qquad (36)$$

Simplifying the inequality in eq. (36) we obtain the constraint on the expert probabilities given in eq. (6).

## 8.2 Appendix B

In this appendix, we prove the bound given in eq. (30):

$$\begin{aligned}
\sum_i \frac{1}{\left(\overline{P}_e P_h^i + P_e \overline{P}_h^i\right)} &\leq \sum_{i: P_h^i \leq 0.5} 2\left(2 - \frac{1}{P_e}\right) P_h^i + \frac{1}{P_e} \\
&\quad + \sum_{i: P_h^i > 0.5} 2\left(\frac{1}{\overline{P}_e} - 2\right) P_h^i + \left(4 - \frac{1}{\overline{P}_e}\right)
\end{aligned} \qquad (37)$$

where $P_h^i = P(Z = y|x, h)$, $\overline{P}_h^i = 1 - P_h^i$, $\overline{P}_e = 1 - P_e$.

*Proof.* Consider the function $f(P_e) = 1/P_e$ defined in the interval $P_e \in [0, 1]$. The function $f$ is convex in the given interval which allows us to bound it using a linear combination of the function values at individual points given by:

$$f(\lambda x_1 + (1 - \lambda) x_2) \leq \lambda f(x_1) + (1 - \lambda) f(x_2) \qquad (38)$$

where $\lambda \in [0, 1]$. To find a bound for $\frac{1}{\left(\overline{P}_e P_h^i + P_e \overline{P}_h^i\right)}$, we consider two different cases for $P_h^i$. In the first instance, we consider $P_h^i$ varying in the interval $[0, 1/2]$ and use the inequality eq. (38) with $\lambda = 2P_h^i$, $x_1 = \left(P_e + \overline{P}_e\right)/2$ and $x_2 = P_e$ to get

$$f\left(2P_h^i \frac{P_e + \overline{P}_e}{2} + \left(1 - 2P_h^i\right) P_e\right) \qquad (39)$$
$$\leq 2P_h^i f\left(\frac{P_e + \overline{P}_e}{2}\right) + \left(1 - 2P_h^i\right) f(P_e)$$

$$\Rightarrow f\left(\overline{P}_e P_h^i + P_e \left(1 - P_h^i\right)\right) \qquad (40)$$
$$\leq 2P_h^i f\left(\frac{P_e + \overline{P}_e}{2}\right) + \left(1 - 2P_h^i\right) f(P_e).$$

Using the definition of $f$ and noting that $\overline{P}_h^i = 1 - P_h^i$ we can rewrite the inequality as

$$\sum_i \frac{1}{\left(\overline{P}_e P_h^i + P_e \overline{P}_h^i\right)} \leq \sum_{i: P_h^i \leq 0.5} 2\left(2 - \frac{1}{P_e}\right) P_h^i + \frac{1}{P_e} \qquad (41)$$

which proves the first part of the inequality eq. (37). We prove the second part when $P_h^i \in [0.5, 1]$ in a similar way. We again use the inequality in eq. (38) with $\lambda = 2P_h^i - 1$, $x_1 = \overline{P}_e$ and $x_2 = 2/\left(P_e + \overline{P}_e\right)$ to obtain:

$$f\left(\left(2P_h^i - 1\right) \overline{P}_e + \left(2 - 2P_h^i\right) \frac{2}{(P_e + \overline{P}_e)}\right) \qquad (42)$$
$$\leq \left(2P_h^i - 1\right) f(\overline{P}_e) + \left(2 - 2P_h^i\right) f\left(\frac{2}{(P_e + \overline{P}_e)}\right)$$

$$\Rightarrow f\left(\overline{P}_e P_h^i + P_e \left(1 - P_h^i\right)\right) \qquad (43)$$
$$\leq \left(2P_h^i - 1\right) f(\overline{P}_e) + \left(2 - 2P_h^i\right) f\left(\frac{2}{(P_e + \overline{P}_e)}\right).$$

Using the definition of $f$ we can rewrite the inequality as:

$$\sum_i \frac{1}{\left(\overline{P}_e P_h^i + P_e \overline{P}_h^i\right)}$$
$$\leq \sum_{i: P_h^i > 0.5} 2\left(\frac{1}{\overline{P}_e} - 2\right) P_h^i + \left(4 - \frac{1}{\overline{P}_e}\right). \qquad (44)$$

Combining eq. (41) and eq. (44) proves the inequality in eq. (37). □